\documentclass[11pt,a4paper]{article}
\usepackage[hyperref]{acl2020}
\usepackage{times}
\usepackage{latexsym}

\usepackage{microtype}

\aclfinalcopy 
\def\aclpaperid{2137} 

\usepackage[]{algorithm2e}
\usepackage{pgfplots}
\usepackage{subcaption}
\usepackage{booktabs}
\usepackage{amssymb}
\pgfplotsset{compat=1.3}

\usepackage{xspace} 
\newcommand{\Rat}{Raw Attention\xspace}
\newcommand{\Jat}{Attention rollout\xspace}
\newcommand{\Fat}{Attention flow\xspace}
\newcommand{\rat}{raw attention\xspace}
\newcommand{\jat}{attention rollout\xspace}
\newcommand{\fat}{attention flow\xspace}
\newcommand{\blankout}{blank-out\xspace}
\newcommand{\Blankout}{Blank-out\xspace}

\title{Quantifying Attention Flow in Transformers}

\author{Samira Abnar \\
 ILLC, University of Amsterdam \\
  \texttt{s.abnar@uva.nl} \\\And
  Willem Zuidema \\
  ILLC, University of Amsterdam  \\
  \texttt{w.h.zuidema@uva.nl} \\}

\date{}

\begin{document}
\maketitle
\begin{abstract}

In the Transformer model, ``self-attention'' combines information from attended embeddings into the representation of the focal embedding in the next layer. Thus, across layers of the Transformer, information originating from different tokens gets increasingly mixed. This makes attention weights unreliable as explanations probes. In this paper, we consider the problem of quantifying this flow of information through self-attention. We propose two methods for approximating the attention to input tokens given attention weights, \emph{\jat} and \emph{\fat}, as post hoc methods when we use attention weights as the relative relevance of the input tokens. We show that these methods give complementary views on the flow of information, and compared to raw attention, both yield higher correlations with importance scores of input tokens obtained using an ablation method and input gradients.
\end{abstract}

\section{Introduction}
Attention~\citep{bahdanau2014neural, vaswani2017attention} has become the key building block of neural sequence processing models, 
and visualizing attention weights is the easiest and most popular approach to interpret a model's decisions and to gain insights about its internals~\citep{vaswani2017attention, xu2015show, wang2016attention,  lee2017interactive, dehghani2018universal, rocktaschel2015reasoning, chen2019improving, coenen2019visualizing, clark-etal-2019-bert}.
Although it is wrong to equate attention with explanation ~\citep{pruthi2019learning,jain2019attention}, it can offer plausible and meaningful interpretations~\citep{wiegreffe2019attention, vashishth2019attention, vig2019visualizing}.
In this paper, we focus on problems arising when we move to the higher layers of a model, due to lack of token identifiability of the embeddings in higher layers~\citep{brunner2019validity}. 

\begin{figure*}
    \centering
    \begin{subfigure}[b]{0.3\textwidth}
         \includegraphics[width=\columnwidth,trim={3.0cm 3.5cm 4.0cm 3.1cm},clip]{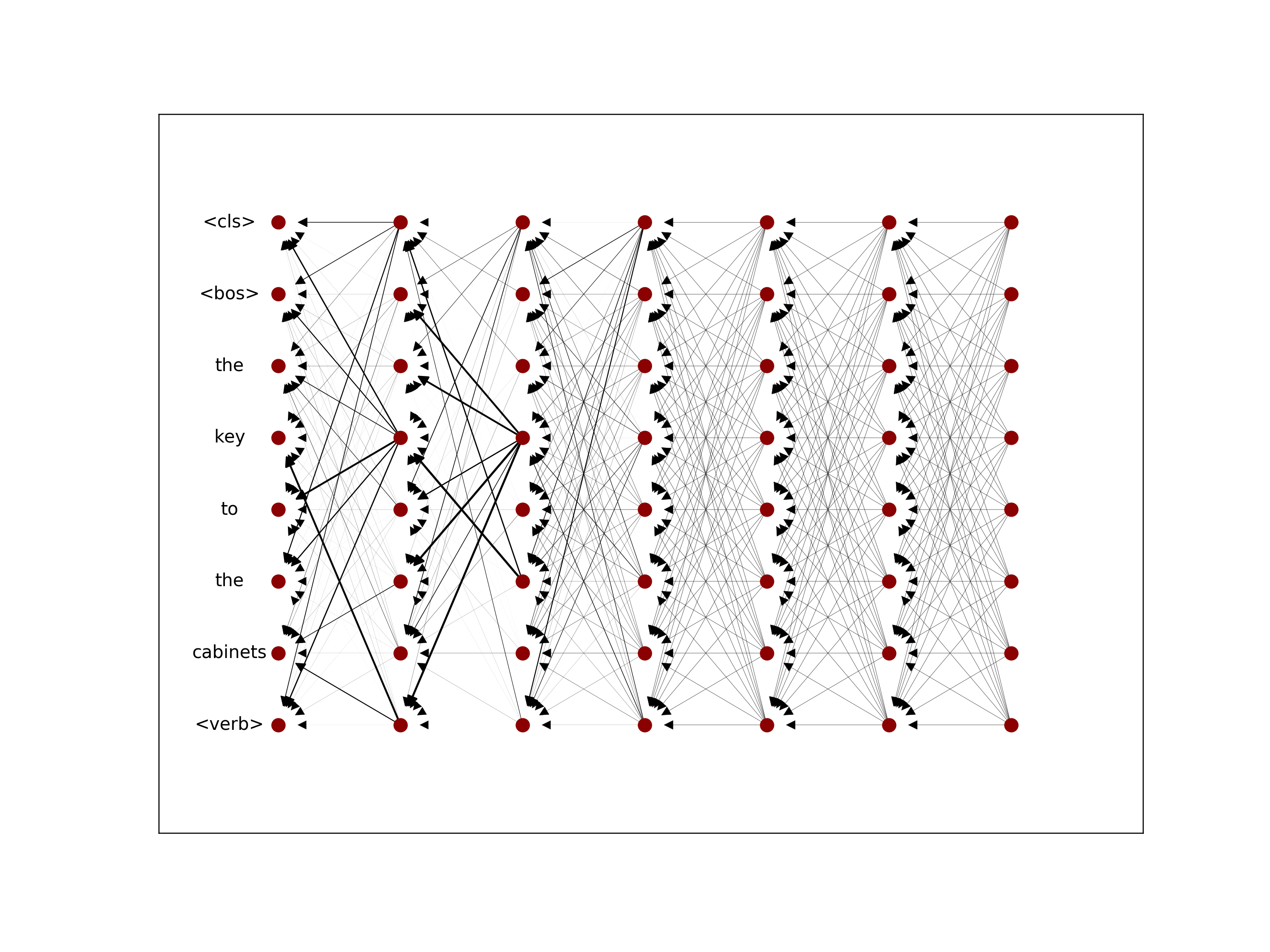}
         \caption{Embedding attentions
         \label{fig:raw_attention}}
     \end{subfigure}%
     \hfill
     \begin{subfigure}[b]{0.3\textwidth}
         \centering
             \includegraphics[width=\columnwidth,trim={3.0cm 3.5cm 3.5cm 3.1cm},clip]{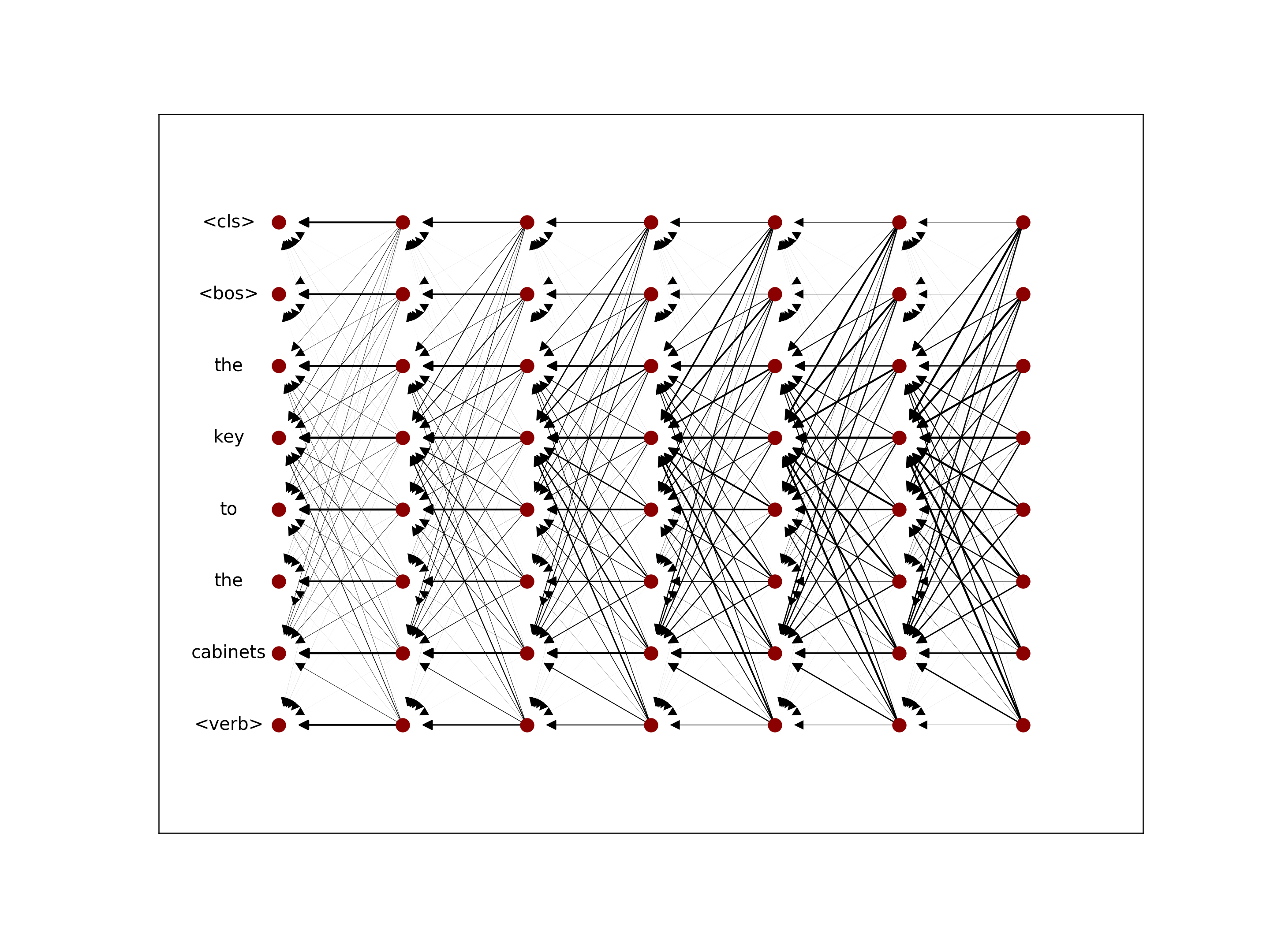}
         \caption{\Jat
         \label{fig:res_joint}}
     \end{subfigure}%
     \hfill
     \begin{subfigure}[b]{0.3\textwidth}
         \centering
         \includegraphics[width=\columnwidth,trim={3.0cm 3.5cm 4.0cm 3.1cm},clip]{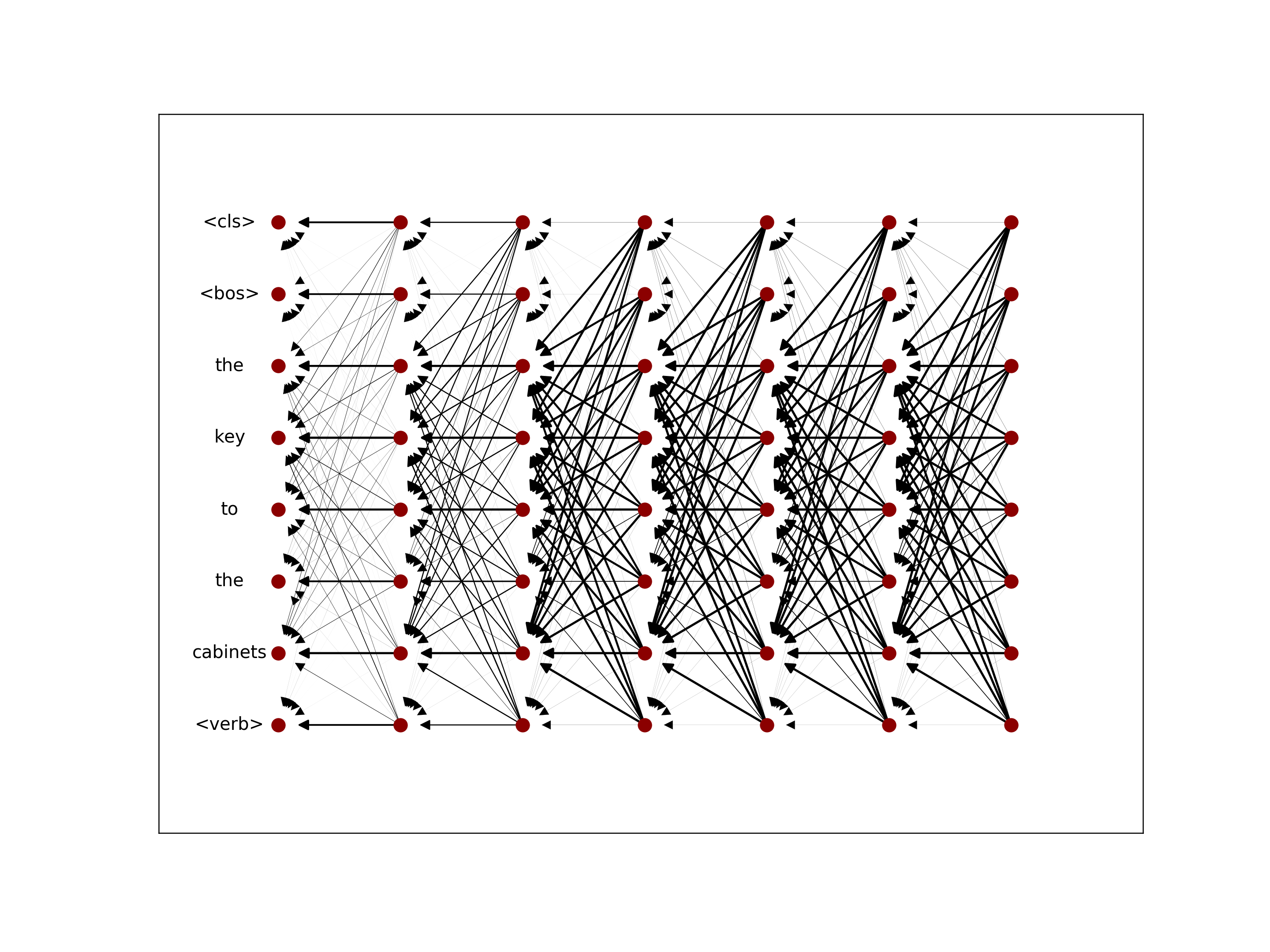}
         \caption{\Fat
         \label{fig:res_flow}}
     \end{subfigure}
    \caption{Visualisation of attention weights.
    \label{fig:att_viz}}
\end{figure*}

\begin{figure}
    \centering
    \begin{subfigure}[b]{0.288\columnwidth}
   \includegraphics[width=\columnwidth, trim={0.0cm 0.0cm 0.0cm 0.0cm},clip]{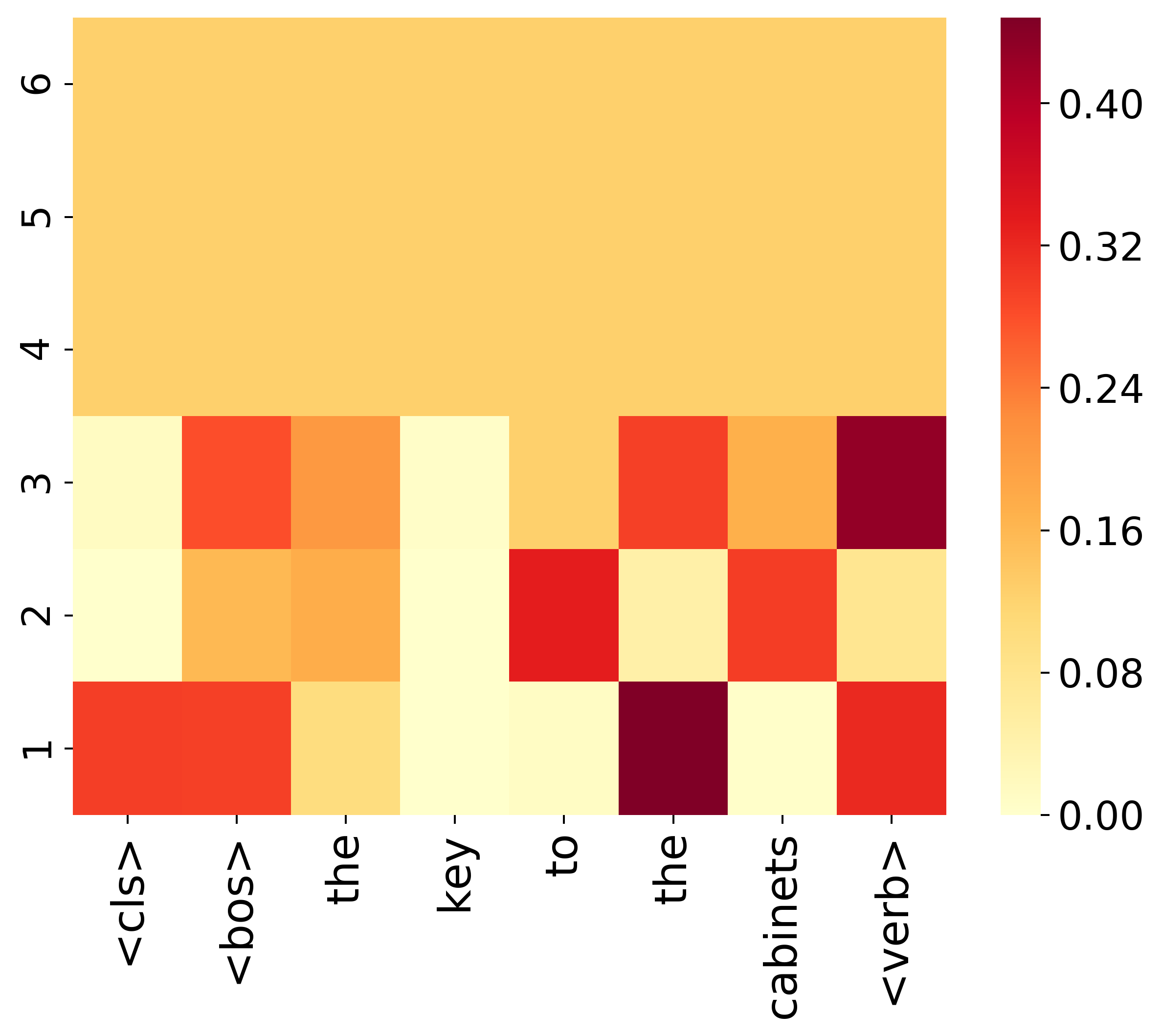}
     \end{subfigure}%
     \begin{subfigure}[b]{0.322\columnwidth}
    \includegraphics[width=\columnwidth, trim={0.0cm 0.0cm 0.0cm 0.0cm},clip]{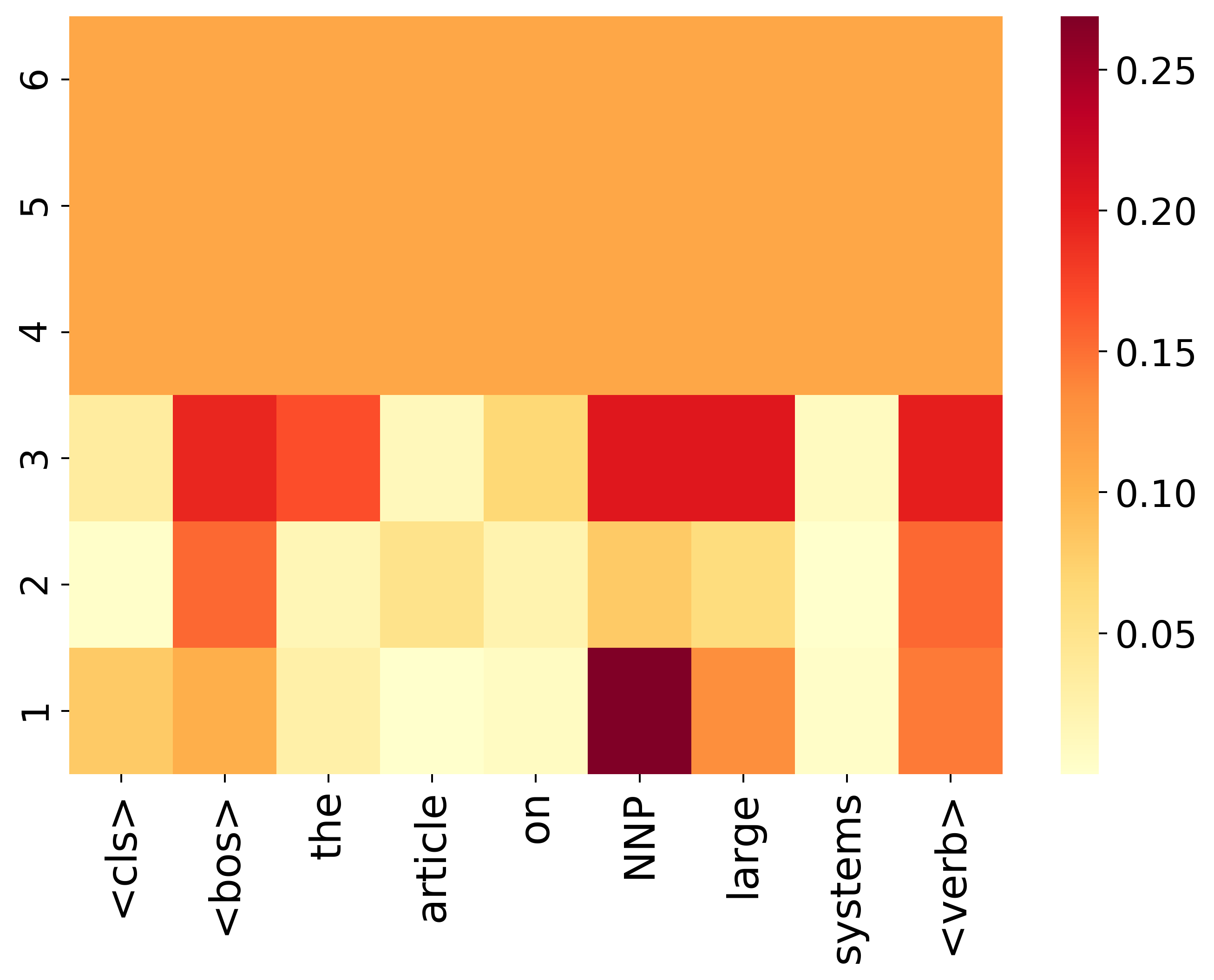}
     \end{subfigure}%
     \begin{subfigure}[b]{0.39\columnwidth}
    \includegraphics[width=\columnwidth, trim={0.0cm 0.0cm 0.0cm 0.0cm},clip]{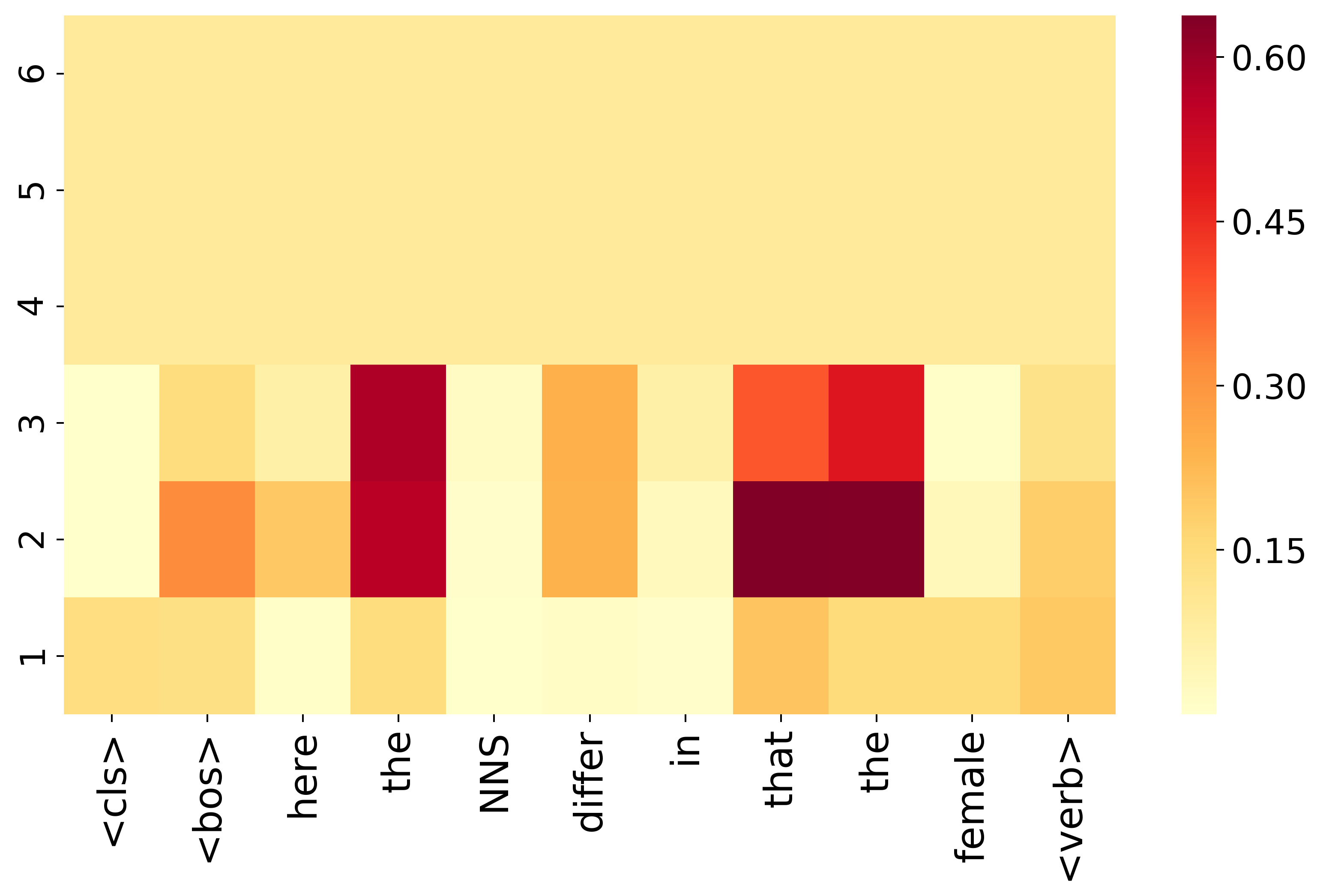}
     \end{subfigure}
    \vspace{-15pt}
    \caption{\Rat maps for the \texttt{CLS} token at different layers.
    \label{fig:rat_cls_maps}}
\vspace{-10pt}
\end{figure}

We propose two simple but effective methods to compute attention scores to input tokens  (i.e., \emph{token attention}) at each layer, by taking raw attentions (i.e., \emph{embedding attention}) of that layer as well as those from the precedent layers. 
These methods are based on modelling the information flow in the network with a \emph{DAG} (Directed Acyclic Graph), in which the nodes are input tokens and hidden embeddings, edges are the attentions from the nodes in each layer to those in the previous layer, and the weights of the edges are the attention weights.
The first method, \emph{\jat}, assumes that the identities of input tokens are linearly combined through the layers based on the attention weights.
To adjust attention weights, it rolls out the weights to capture the propagation of information from input tokens to intermediate hidden embeddings.
The second method, \emph{\fat}, considers the attention graph as a flow network. Using a maximum flow algorithm, it computes maximum flow values, from hidden embeddings (sources) to input tokens (sinks).
In both methods, we take the residual connection in the network into account to better model the connections between input tokens and hidden embedding. 
We show that compared to \rat, the token attentions from \jat and \fat have higher correlations with the importance scores obtained from input gradients as well as \emph{\blankout}, an input ablation based attribution method.
Furthermore, we visualize the token attention weights and demonstrate that they are better approximations of how input tokens contribute to a predicted output, compared to \rat. 

It is noteworthy that the techniques we propose in this paper, are not toward making hidden embeddings more identifiable, or providing better attention weights for better performance, but a new set of attention weights that take token identity problem into consideration and can serve as a better diagnostic tool for visualization and debugging.

\section{Setups and Problem Statement}
In our analysis, we focus on the verb number prediction task, i.e., predicting singularity or plurality of a verb of a sentence, when the input is the sentence up to the verb position. We use the subject-verb agreement dataset~\citep{linzen2016assessing}. 
This task and dataset are convenient choices, as they offer a clear hypothesis about what part of the input is essential to get the right solution. For instance, given ``\emph{the key to the cabinets}'' as the input, we know that attending to ``key'' helps the model predict singular as output while attending to ``cabinets'' (an \emph{agreement attractor}, with the opposite number) is unhelpful.

We train a Transformer encoder, with GPT-2 Transformer blocks as described in~\citep{radford2019language, wolf2019huggingfacests} (without masking). The model has 6 layers, and 8 heads, with hidden/embedding size of 128. 
Similar to Bert~\citep{devlin2018bert} we add a \texttt{CLS} token and use its embedding in the final layer as the input to the classifier. The accuracy of the model on the subject-verb agreement task is $0.96$.
To facilitate replication of our experiments we will make the implementations of the models we use and algorithms we introduce  publicly available at \url{https://github.com/samiraabnar/attention_flow}.

We start by visualizing \rat in Figure~\ref{fig:raw_attention} (like~\citealt{vig2019visualizing}). The example given here is correctly classified. Crucially, only in the first couple of layers, there are some distinctions in the attention patterns for different positions, while in higher layers the attention weights are rather uniform. Figure~\ref{fig:rat_cls_maps} (left) gives raw attention scores of the \texttt{CLS} token over input tokens (x-axis) at  different layers (y-axis), which similarly lack an interpretable pattern.
These observations reflect the fact that as we go deeper into the model, the embeddings are more contextualized and may all carry similar information. 
This underscores the need to track down attention weights all the way back to the input layer and is in line with findings of \citet{serrano2019attention}, who show that attention weights do not necessarily correspond to the relative importance of input tokens.  

To quantify the usefulness of raw attention weights, and the two alternatives that we consider in the next section, 
besides input gradients, we employ an input ablation method, \emph{\blankout}, to estimate an importance score for each input token. \Blankout replaces each token in the input, one by one, with \texttt{UNK} and measures how much it affects the predicted probability of the correct class.
We compute the \emph{Spearman's rank correlation} coefficient between the attention weights of the \texttt{CLS} embedding in the final layer and the importance scores from \blankout. 
As shown in the first row of Table~\ref{tab:quant_comp_blank}, the correlation between \rat weights of the \texttt{CLS} token and \blankout scores is rather low, except for the first layer. As we can see in Table~\ref{tab:quant_comp_grad} this is also the case when we compute the correlations with input gradients. 

\begin{table}[!ht]
    \centering
    \resizebox{\columnwidth}{!}{%
    \begin{tabular}{lcccccc}
        \toprule
          & L1 & L2 & L3 & L4 & L5 & L6 \\
         \midrule
         Raw & 0.69$\pm$0.27 & 0.10$\pm$0.43 & -0.11$\pm$0.49 & -0.09$\pm$0.52 & 0.20$\pm$0.45 & 0.29$\pm$0.39 \\
         Rollout & 0.32$\pm$0.26 & 0.38$\pm$0.27 & 0.51$\pm$0.26 & 0.62$\pm$0.26 & 0.70$\pm$0.25 & 0.71$\pm$0.24 \\
         Flow & 0.32$\pm$0.26 & 0.44$\pm$0.29 & 0.70$\pm$0.25 & 0.70$\pm$0.22 & 0.71$\pm$0.22 & 0.70$\pm$0.22 \\
         \bottomrule
    \end{tabular}
    }
    \caption{SpearmanR correlation of attention based importance with \blankout scores for 2000 samples from the test set for the verb number prediction model.}
    \label{tab:quant_comp_blank}
\end{table}

\begin{table}[!ht]
    \centering
    \resizebox{\columnwidth}{!}{%
    \begin{tabular}{lcccccc}
        \toprule
          & L1 & L2 & L3 & L4 & L5 & L6 \\
         \midrule
         Raw &  0.53$\pm$0.33 & 0.16$\pm$0.38 & -0.06$\pm$0.42 & 0.00$\pm$0.47 & 0.24$\pm$0.40 & 0.46$\pm$0.35 \\
         Rollout &  0.22$\pm$0.31 & 0.27$\pm$0.32 & 0.39$\pm$0.32 & 0.47$\pm$0.32 & 0.53$\pm$0.32 & 0.54$\pm$0.31 \\
         Flow &  0.22$\pm$0.31 & 0.31$\pm$0.34 & 0.54$\pm$0.32 & 0.61$\pm$0.28 & 0.60$\pm$0.28 & 0.61$\pm$0.28 \\
         \bottomrule
    \end{tabular}
    }
    \caption{SpearmanR correlation of attention based importance with input gradients for 2000 samples  from the test set for the verb number prediction model.}
    \label{tab:quant_comp_grad}
\end{table}

\section{Attention Rollout and Attention Flow}
\Jat and \fat recursively compute the token attentions in each layer of a given model given the embedding attentions as input. They differ in the assumptions they make about how attention weights in lower layers affect the flow of information to the higher layers and whether to compute the token attentions relative to each other or independently. 

To compute how information propagates from the input layer to the embeddings in higher layers, it is crucial to take the residual connections in the model into account as well as the attention weights. 
In a Transformer block, both self-attention and feed-forward networks are wrapped by residual connections, i.e., the input to these modules is added to their output. 
When we only use attention weights to approximate the flow of information in Transformers, we ignore the residual connections. But these connections play a significant role in tying corresponding positions in different layers. 
Hence, to compute \jat and \fat, we augment the attention graph with extra weights to represent residual connections.
Given the attention module with residual connection, we compute values in layer $l+1$ as $V_{l+1} = V_{l}  + W_{att}V_l$, where $ W_{att}$ is the attention matrix. Thus, we have $V_{l+1} = (W_{att} + I) V_{l}$. So, to account for residual connections, we add an identity matrix to the attention matrix and re-normalize the weights. This results in $A = 0.5W_{att} + 0.5I$, where $A$ is the raw attention updated by residual connections. 

Furthermore, analyzing individual heads requires accounting for mixing of information between heads through a position-wise feed-forward network in Transformer block. Using \jat and \fat, it is also possible to analyze each head separately. We explain in more details in Appendix~\ref{app:singlehead}. However, in our analysis in this paper, for simplicity, we average the attention at each layer over all heads. 

\paragraph{\Jat}
\Jat is an intuitive way of tracking down the information propagated from the input layer to the embeddings in the higher layers.
Given a Transformer with $L$ layers, we want to compute the attention from all positions in layer $l_i$ to all positions in layer $l_j$, where $j<i$.
In the attention graph, a path from node $v$ at position $k$ in $l_i$, to node $u$ at position $m$ in $l_j$, is a series of edges that connect these two nodes. 
If we look at the weight of each edge as the proportion of information transferred between two nodes, we can compute how much of the information at $v$ is propagated to $u$ through a particular path by multiplying the weights of all edges in that path.
Since there may be more than one path between two nodes in the attention graph, to compute the total amount of information propagated from $v$ to $u$, we sum over all possible paths between these two nodes.
At the implementation level, to compute the attentions from $l_i$ to $l_j$, we recursively multiply the attention weights matrices in all the layers below. 
\begin{equation}
  \tilde{A}(l_i)=\left\{
  \begin{array}{@{}ll@{}}
    A(l_i)\tilde{A}(l_{i-1}) & \mbox{if}~ i>j \\
    A(l_i) & \mbox{if}~ i=j
  \end{array}\right.
\label{eq:joint_attention}
\end{equation} 
In this equation, $\tilde{A}$ is \jat, $A$ is \rat and the multiplication operation is a matrix multiplication. With this formulation, to compute input attention we set $j=0$.

\paragraph{\Fat}
In graph theory, a flow network is a directed graph with a ``capacity'' associated with each edge.
Formally, given $G = (V,E)$ is a graph, where $V$ is the set of nodes, and $E$ is the set of edges in $G$;
{$C = \{c_{uv} \in \mathbb{R} ~|~ \forall{u,v} ~\mathrm{where}~ e_{u,v} \in E \land u \ne v \}$} denotes the capacities of the edges and $s,t \in V$ are the source and target (sink) nodes respectively; \emph{flow} is a mapping of edges to real numbers, $f: E \to \mathbb{R}$, that satisfies two conditions: 
(a) \emph{capacity constraint}: for each edge the flow value should not exceed its capacity, $|f_{uv} \leq c_{uv}|$; 
(b) \emph{flow conservation}: for all nodes except $s$ and $t$ the input flow should be equal to output flow --sum of the flow of outgoing edges should be equal to sum of the flow of incoming edges.
Given a flow network, a maximum flow algorithm finds a flow which has the maximum possible value between $s$ and $t$~\citep{clrs}.

Treating the attention graph as a flow network, where the capacities of the edges are attention weights, using any maximum flow algorithm, we can compute the maximum attention flow from any node in any of the layers to any of the input nodes. We can use this maximum-flow-value as an approximation of the attention to input nodes. 
In \fat, the weight of a single path is the minimum value of the weights of the edges in the path, instead of the product of the weights. Besides, we can not compute the attention for node $s$ to node $t$  by adding up the weights of all paths between these two nodes, since there might be an overlap between the paths and this might result in overflow in the overlapping edges.

It is noteworthy that both of the proposed methods can be computed in polynomial time. $O(d*n^2)$ for \jat and $O(d^2*n^4)$ for \fat, where $d$ is the depth of the model, and $n$ is the number of tokens.

\begin{figure}
    \centering
    \begin{subfigure}[b]{\columnwidth}
\begin{tikzpicture}
    \node[anchor=south west,inner sep=0] at (0,0) {\includegraphics[width=0.28\columnwidth, trim={0.0cm 0.0cm 0.0cm 0.0cm},clip]{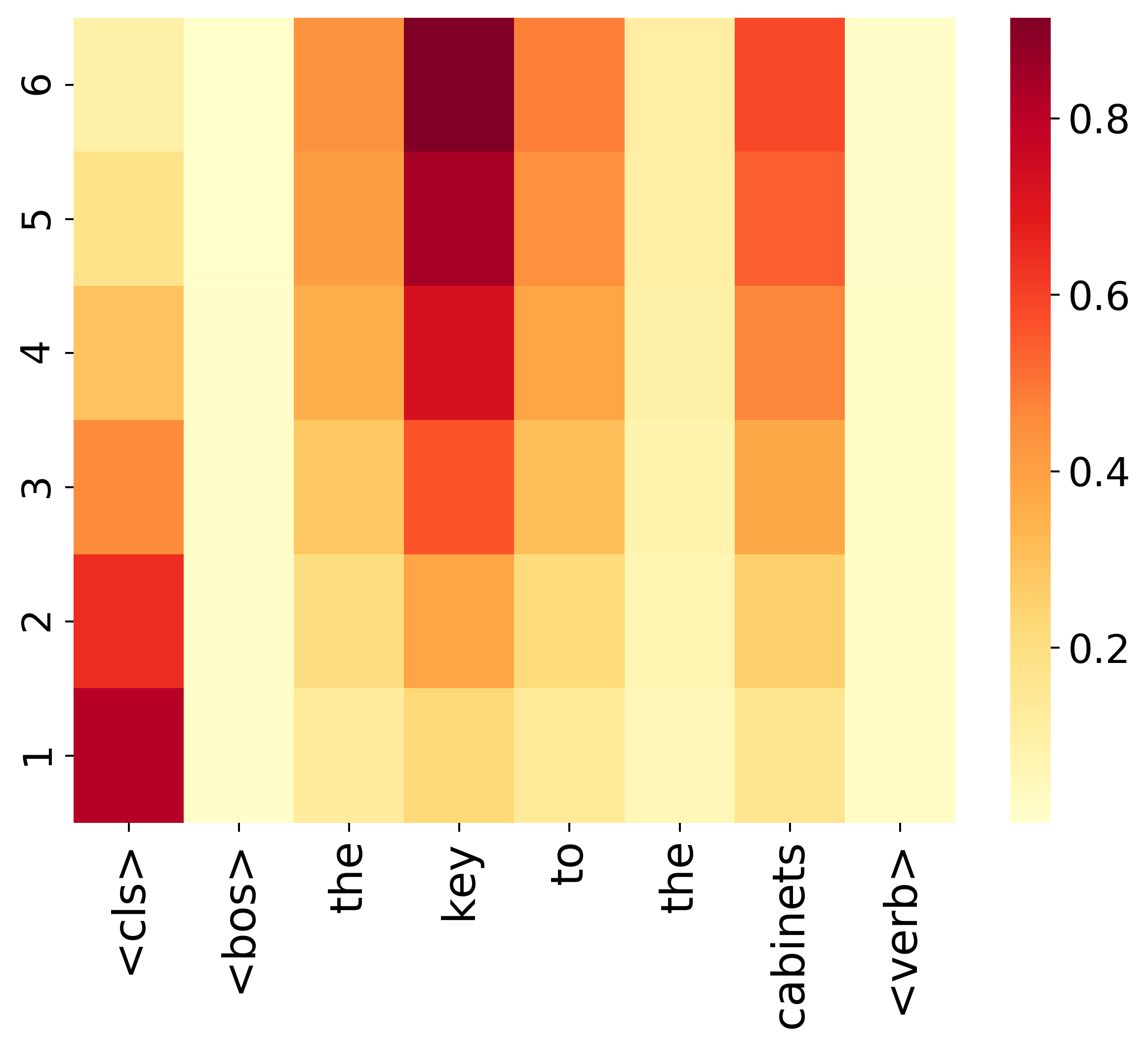}};
    \node[rotate=90] at (-0.1,1.15) {\tiny
{\jat}};  
\end{tikzpicture}%
\begin{tikzpicture}
    \node[anchor=south west,inner sep=0] at (0,0) {\includegraphics[width=0.32\columnwidth, trim={0.0cm 0.0cm 0.0cm 0.0cm},clip]{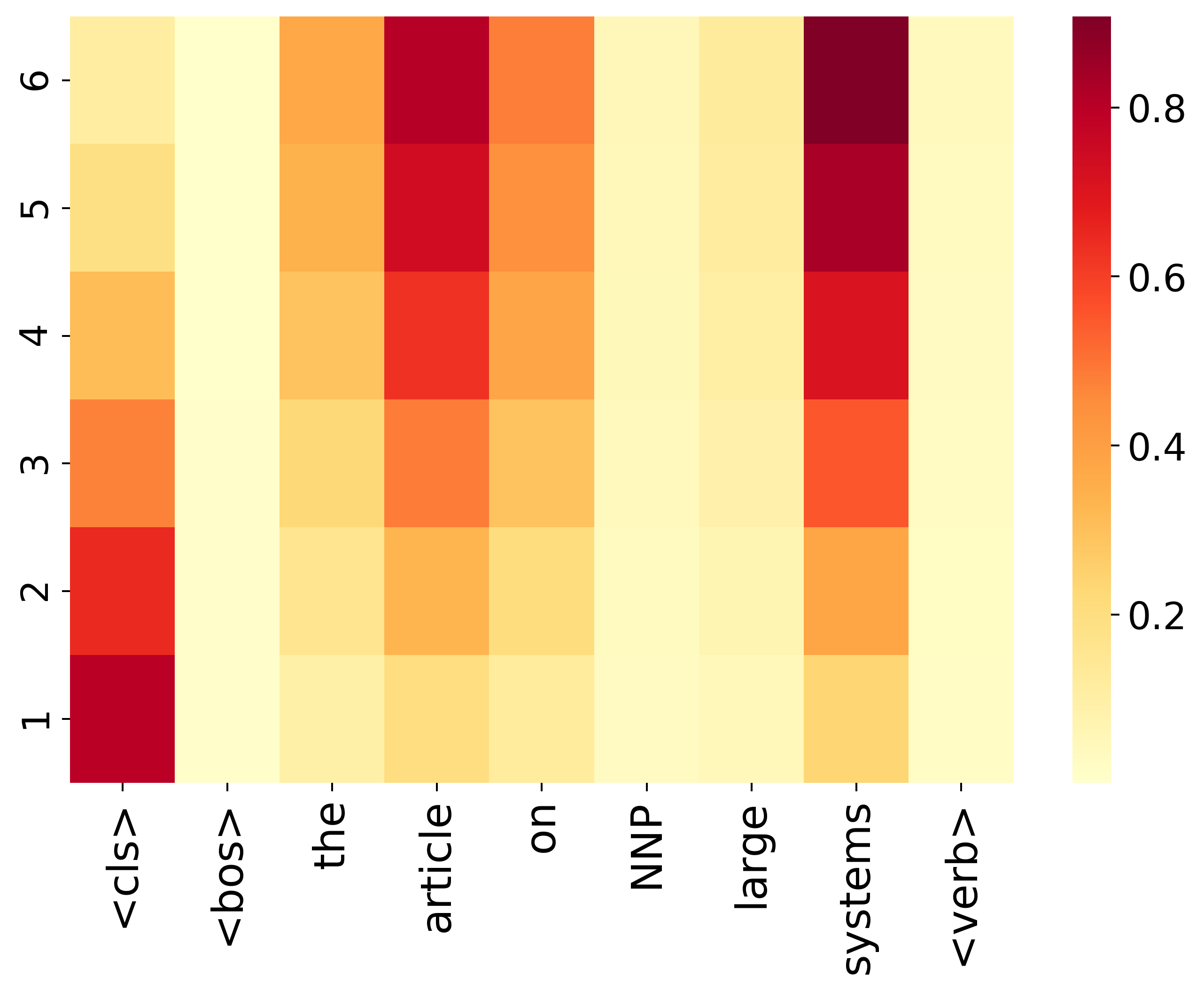}};
\end{tikzpicture}%
\begin{tikzpicture}
    \node[anchor=south west,inner sep=0] at (0,0) {\includegraphics[width=0.39\columnwidth, trim={0.0cm 0.0cm 0.0cm 0.0cm},clip]{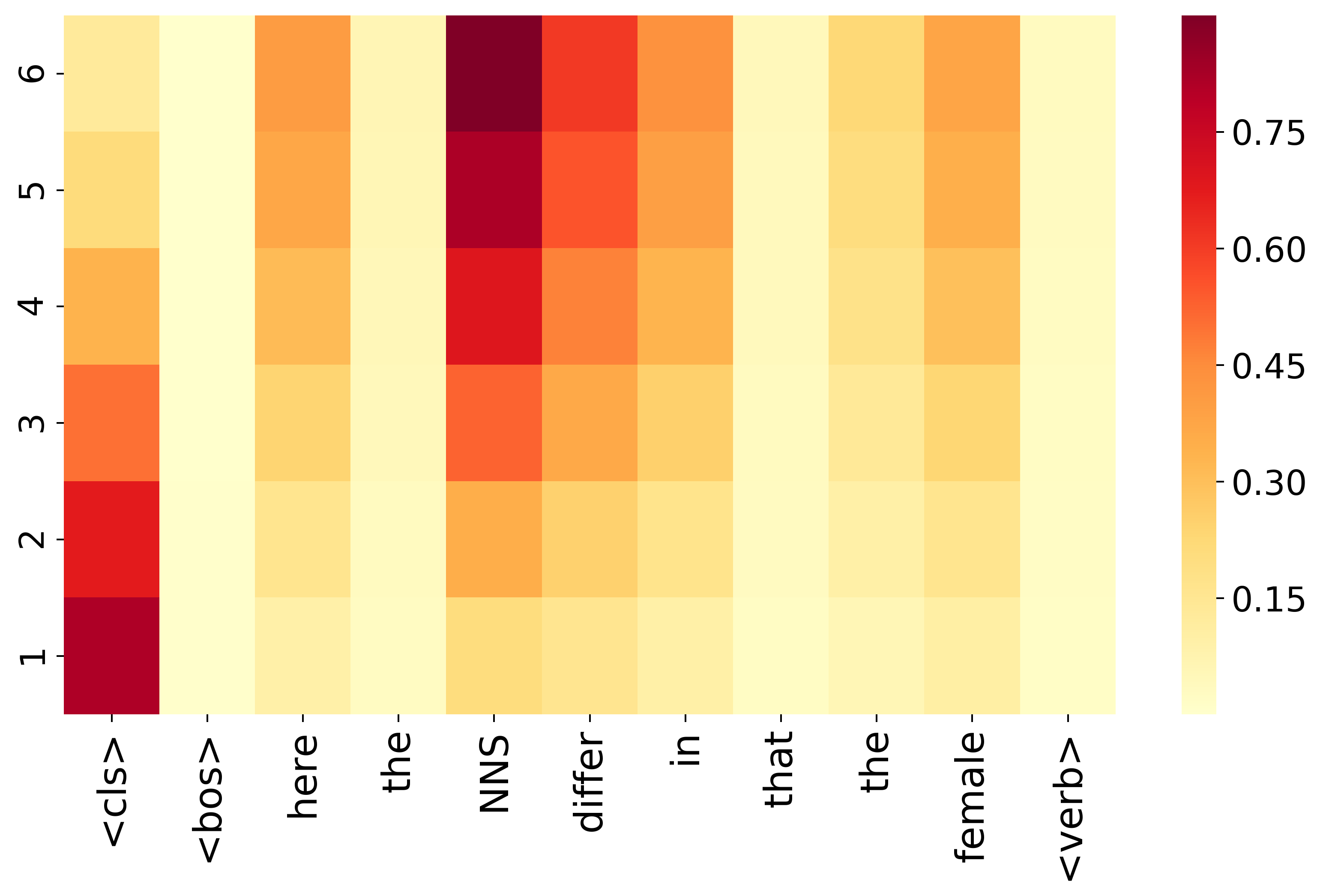}};
\end{tikzpicture}
     \end{subfigure}
    \begin{subfigure}[b]{\columnwidth}
\begin{tikzpicture}
    \node[anchor=south west,inner sep=0] at (0,0) {\includegraphics[width=0.28\columnwidth, trim={0.0cm 0.0cm 0.0cm 0.0cm},clip]{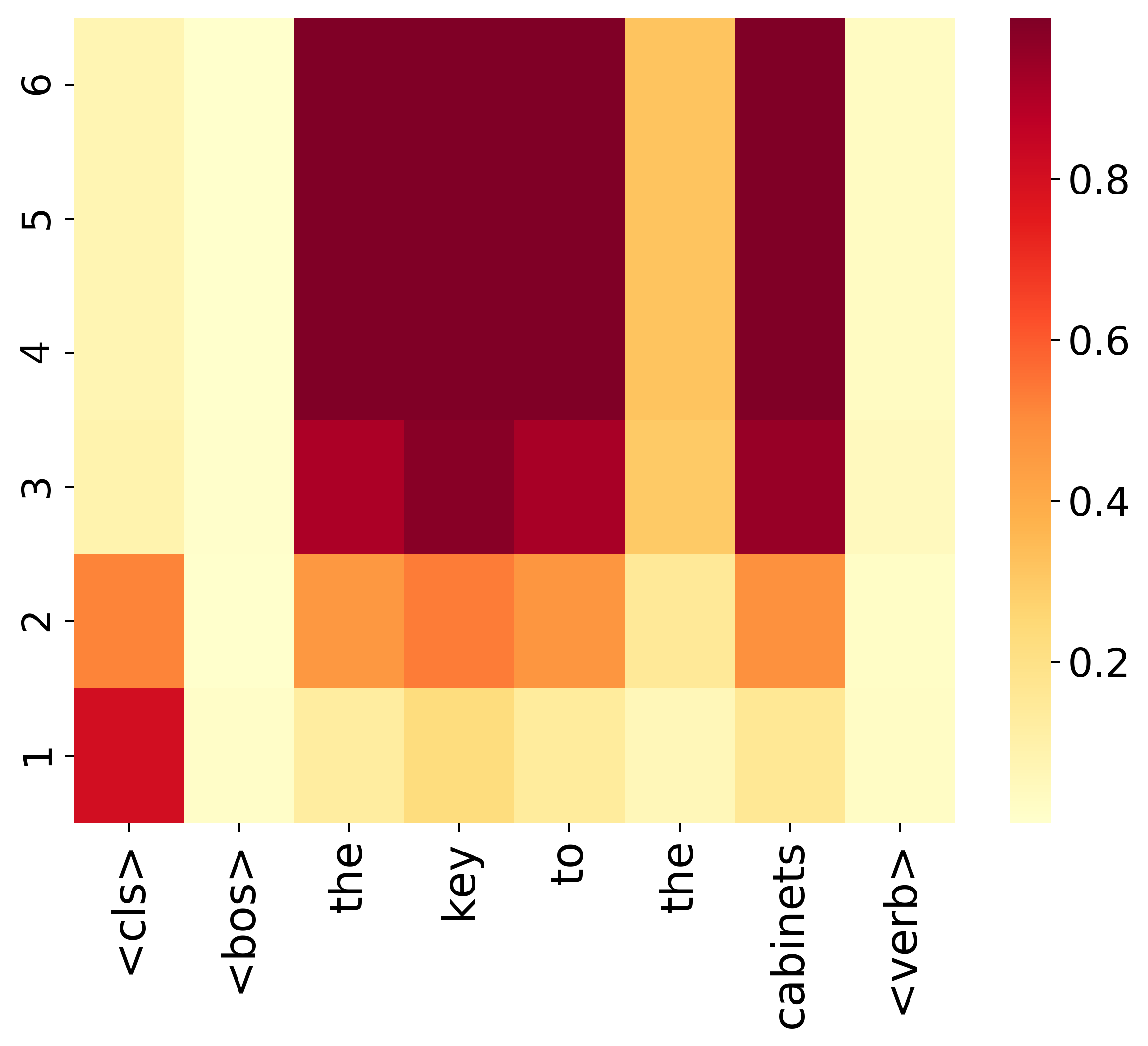}};
    \node[rotate=90] at  (-0.1,1.2) {\tiny
{\fat}};
\end{tikzpicture}%
\begin{tikzpicture}
    \node[anchor=south west,inner sep=0] at (0,0) {\includegraphics[width=0.32\columnwidth, trim={0.0cm 0.0cm 0.0cm 0.0cm},clip]{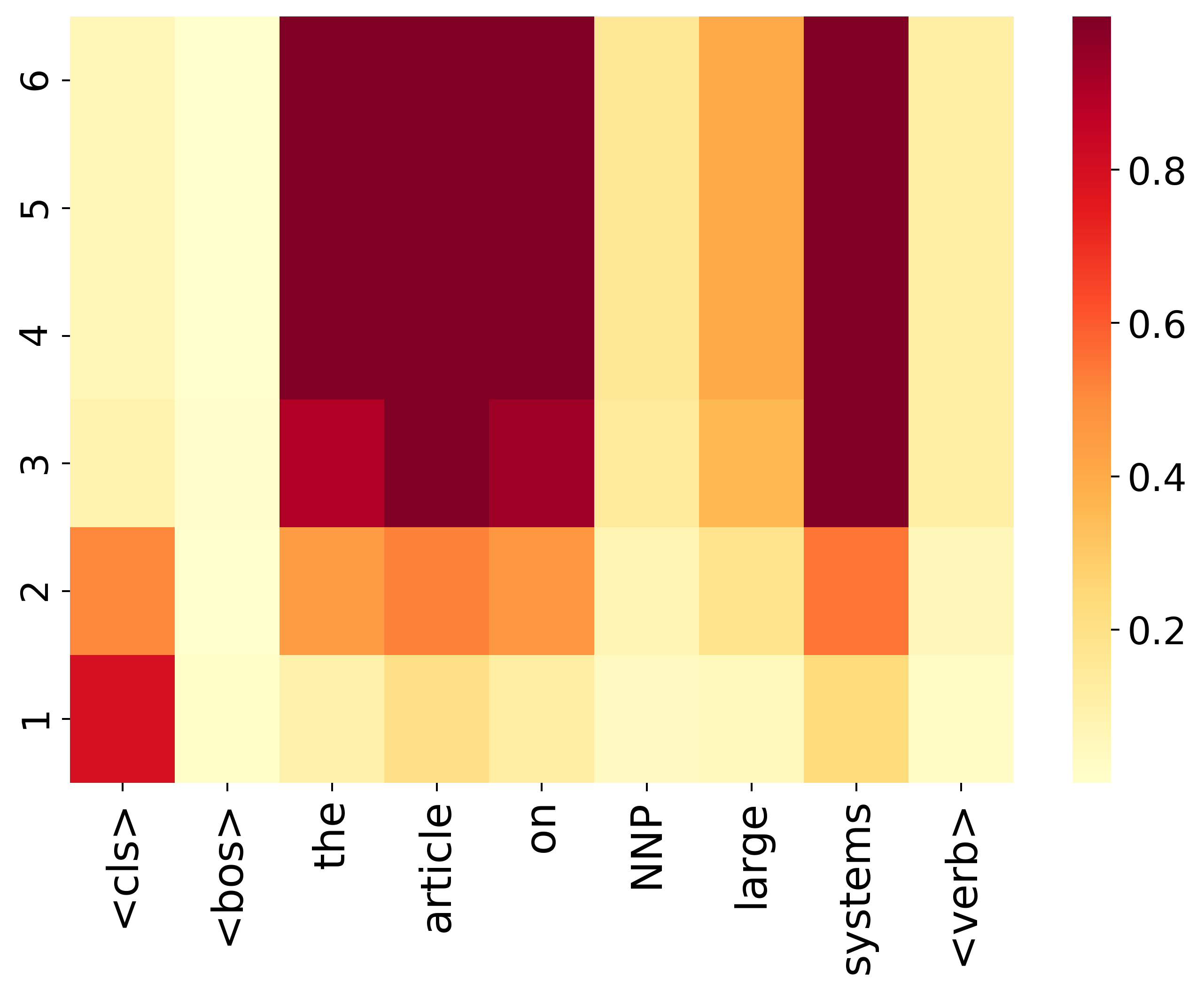}};
\end{tikzpicture}%
\begin{tikzpicture}
    \node[anchor=south west,inner sep=0] at (0,0) {\includegraphics[width=0.39\columnwidth, trim={0.0cm 0.0cm 0.0cm 0.2cm},clip]{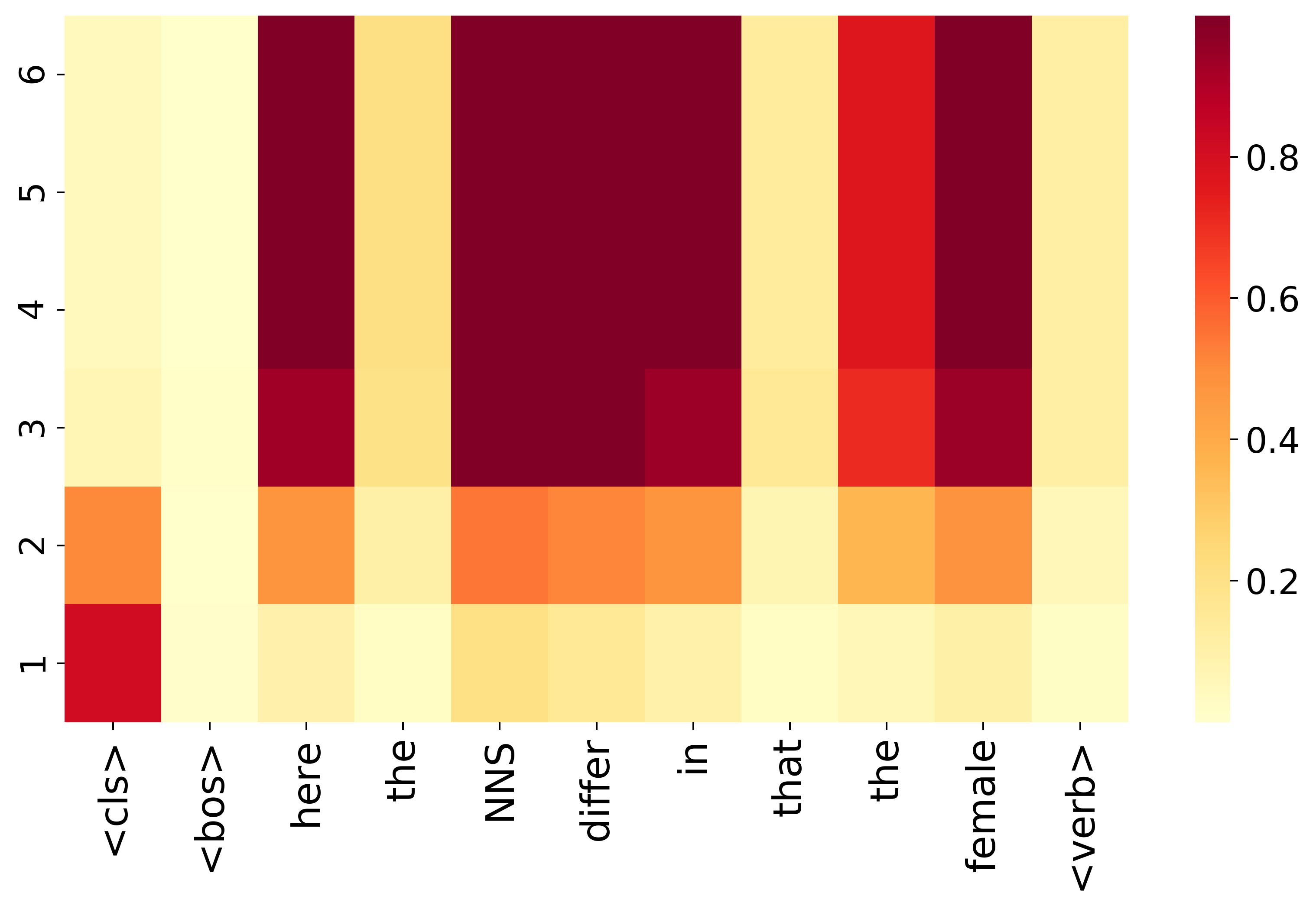}};
\end{tikzpicture}
     \end{subfigure}
    \vspace{-15pt}
    \caption{Attention maps for the \texttt{CLS} token}.
    \label{fig:cls_maps}
\vspace{-17pt}
\end{figure}

\section{Analysis and Discussion}
Now, we take a closer look at these three views of attention.
Figure~\ref{fig:att_viz} depicts \rat, \jat and \fat for a correctly classified example across different layers. It is noteworthy that the first layer of \jat and \fat are the same, and their only difference with \rat is the addition of residual connections. As we move to the higher layers, we see that the residual connections fade away. Moreover, in contrast to \rat, the patterns of \jat and \fat become more distinctive in the higher layers.

Figures~\ref{fig:rat_cls_maps} and \ref{fig:cls_maps} show the weights from \rat, \jat and \fat for the \texttt{CLS} embedding over input tokens (x-axis) in all 6 layers (y-axis) for three examples.
The first example is the same as the one in Figure~\ref{fig:att_viz}.
The second example is ``\emph{the article on NNP large systems} \texttt{<?>}''. The model correctly classifies this example and changing the subject of the missing verb from ``article'' to ``article\textbf{s}'' flips the decision of the model. 
%
The third example is ``\emph{here the NNS differ in that the female} \texttt{<?>}'', which is a miss-classified example and again changing ``NNS'' (plural noun) to ``NNP'' (singular proper noun) flips the decision of the model.

For all cases, the \rat weights are almost uniform above layer three (discussed before). In the case of the correctly classified example, we observe that both \jat and \fat assign relatively high weights to both the subject of the verb, ``article' and the attractor, ``systems''. For the miss-classified example, both \jat and \fat assign relatively high scores to the ``NNS'' token which is not the subject of the verb. This can explain the wrong prediction of the model.

The main difference between \jat and \fat is that \fat weights are amortized among the set of most attended tokens, as expected. \Fat can indicate a set of input tokens that are important for the final decision. Thus we do not get sharp distinctions among them. On the other hand, \jat weights are more focused compared to \fat weights, which is sensible for the third example but not as much for the second one.

\begin{table}[!ht]
    \centering
    \resizebox{\columnwidth}{!}{%
    \begin{tabular}{lcccccc}
        \toprule
          & L1 & L3 & L5 & L6 \\
         \midrule
         Raw &  0.12 $\pm$ 0.21 & 0.09 $\pm$ 0.21 & 0.08 $\pm$ 0.20 & 0.09 $\pm$ 0.21\\
         Rollout & 0.11 $\pm$ 0.19 & 0.12 $\pm$ 0.21 & 0.13 $\pm$ 0.21 & 0.13 $\pm$ 0.20 \\
         Flow & 0.11 $\pm$ 0.19 &  0.11 $\pm$ 0.21 & 0.12 $\pm$ 0.22 & 0.14 $\pm$ 0.21\\
         \bottomrule
    \end{tabular}
    }
    \caption{SpearmanR correlation of attention based importance with input gradients for 100 samples from the test set for the DistillBERT model fine tuned on SST-2.}
    \label{tab:bertsst_quant}
\end{table}








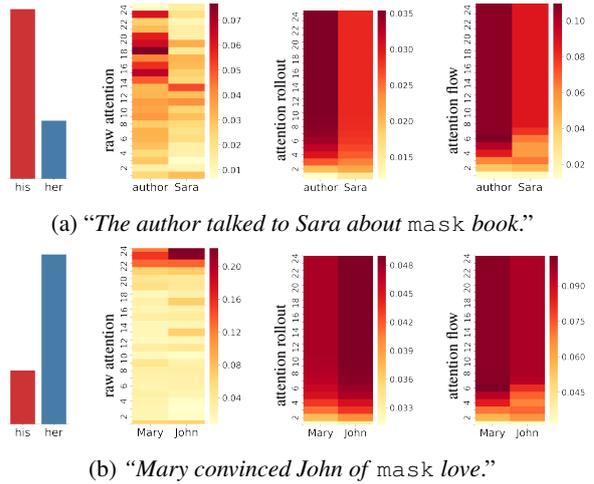
\begin{figure}
    \centering
    \begin{subfigure}[b]{\columnwidth}
     \begin{tikzpicture}
    \node[anchor=south west,inner sep=0] at (0,0)
    {\includegraphics[width=0.118\columnwidth, trim={0.0cm 0.0cm 0.0cm 0.0cm},clip]{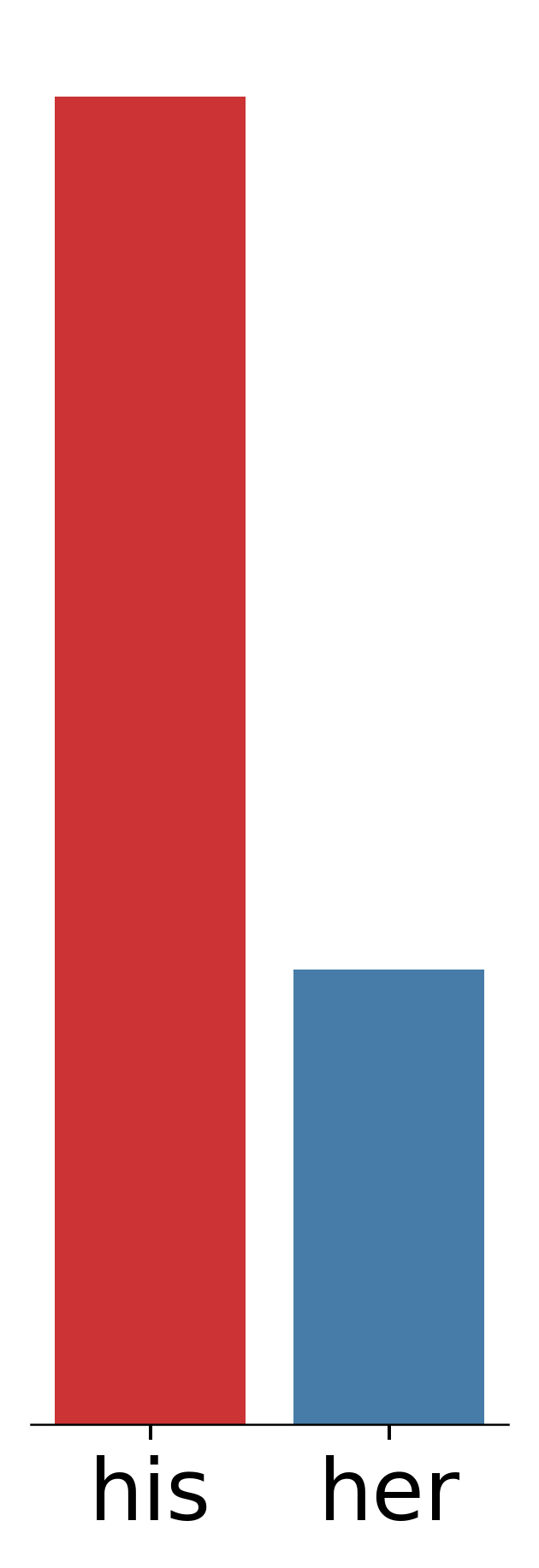}};
    \end{tikzpicture}
         \hfill
    \begin{tikzpicture}
    \node[anchor=south west,inner sep=0] at (0,0)
    {\includegraphics[width=0.22\columnwidth, trim={0.0cm 0.0cm 0.0cm 0.0cm},clip]{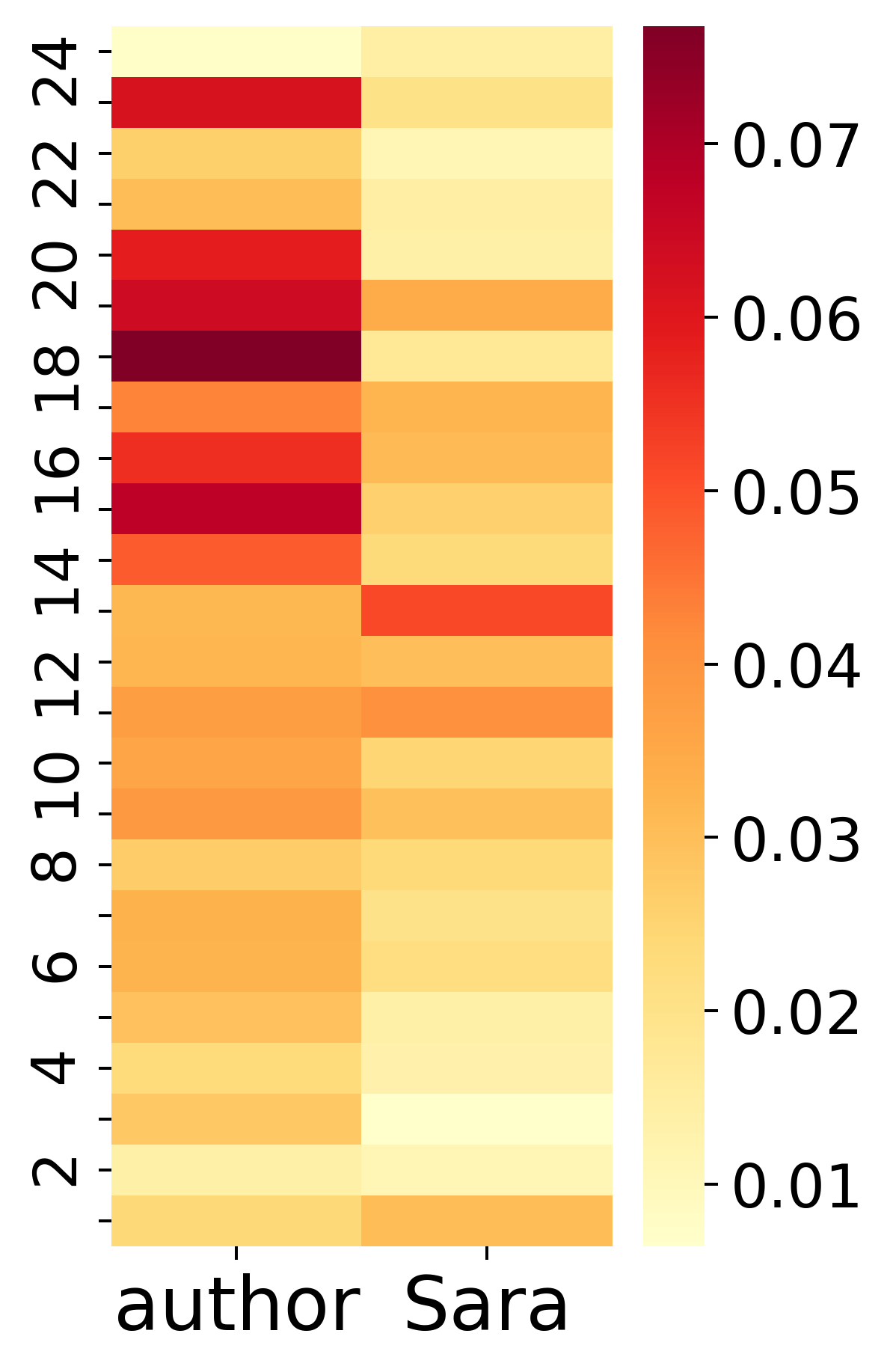}};
    \node[rotate=90] at (-0.1,1.15) {\tiny{\rat}};
     \end{tikzpicture}
          \hfill
    \begin{tikzpicture}
    \node[anchor=south west,inner sep=0] at (0,0)
    {\includegraphics[width=0.22\columnwidth, trim={0.0cm 0.0cm 0.0cm 0.0cm},clip]{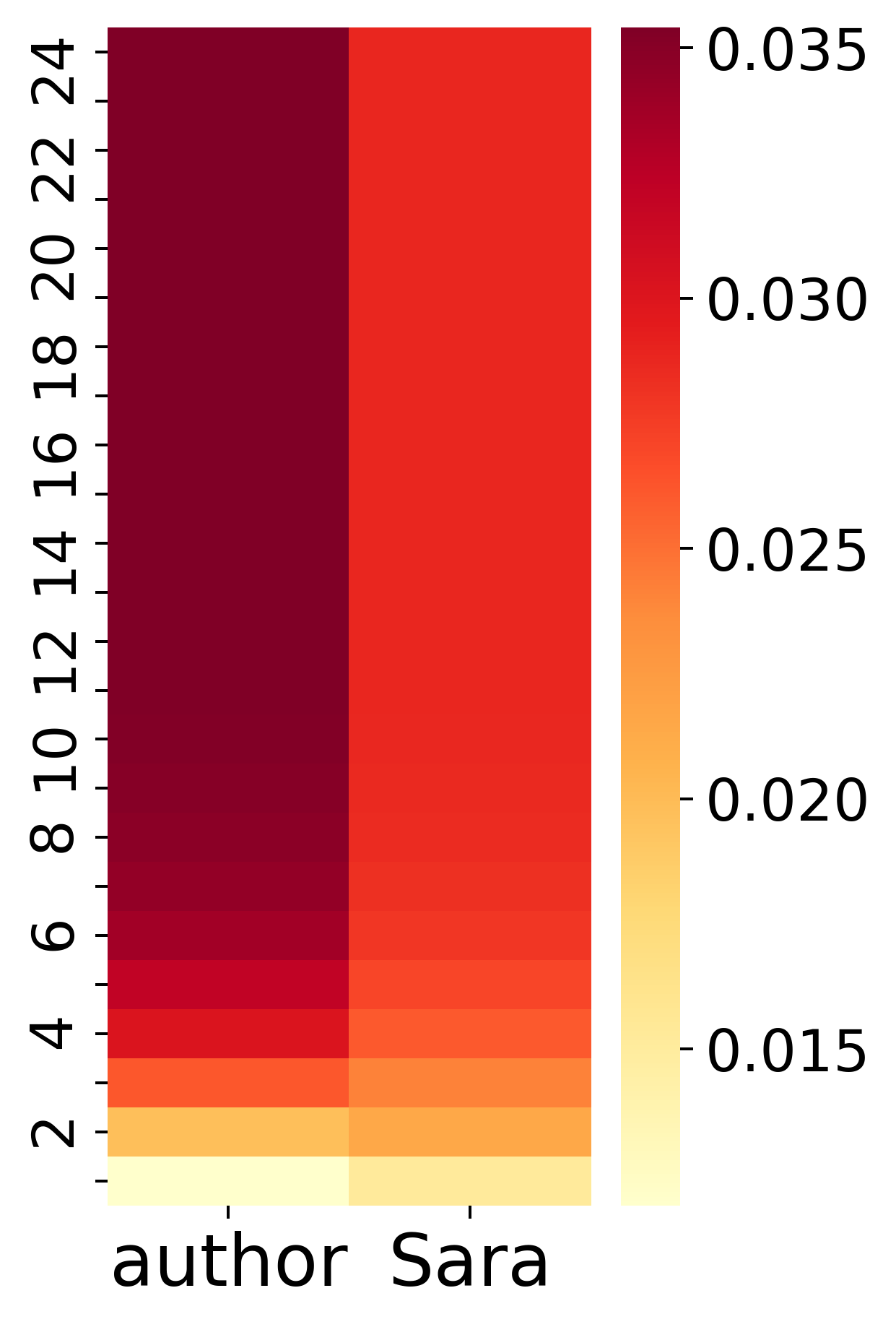}};
    \node[rotate=90] at (-0.1,1.15) {\tiny{\jat}};
     \end{tikzpicture}
          \hfill
    \begin{tikzpicture}
    \node[anchor=south west,inner sep=0] at (0,0)
    {\includegraphics[width=0.22\columnwidth,trim={0.0cm 0.0cm 0.0cm 0.0cm},clip]{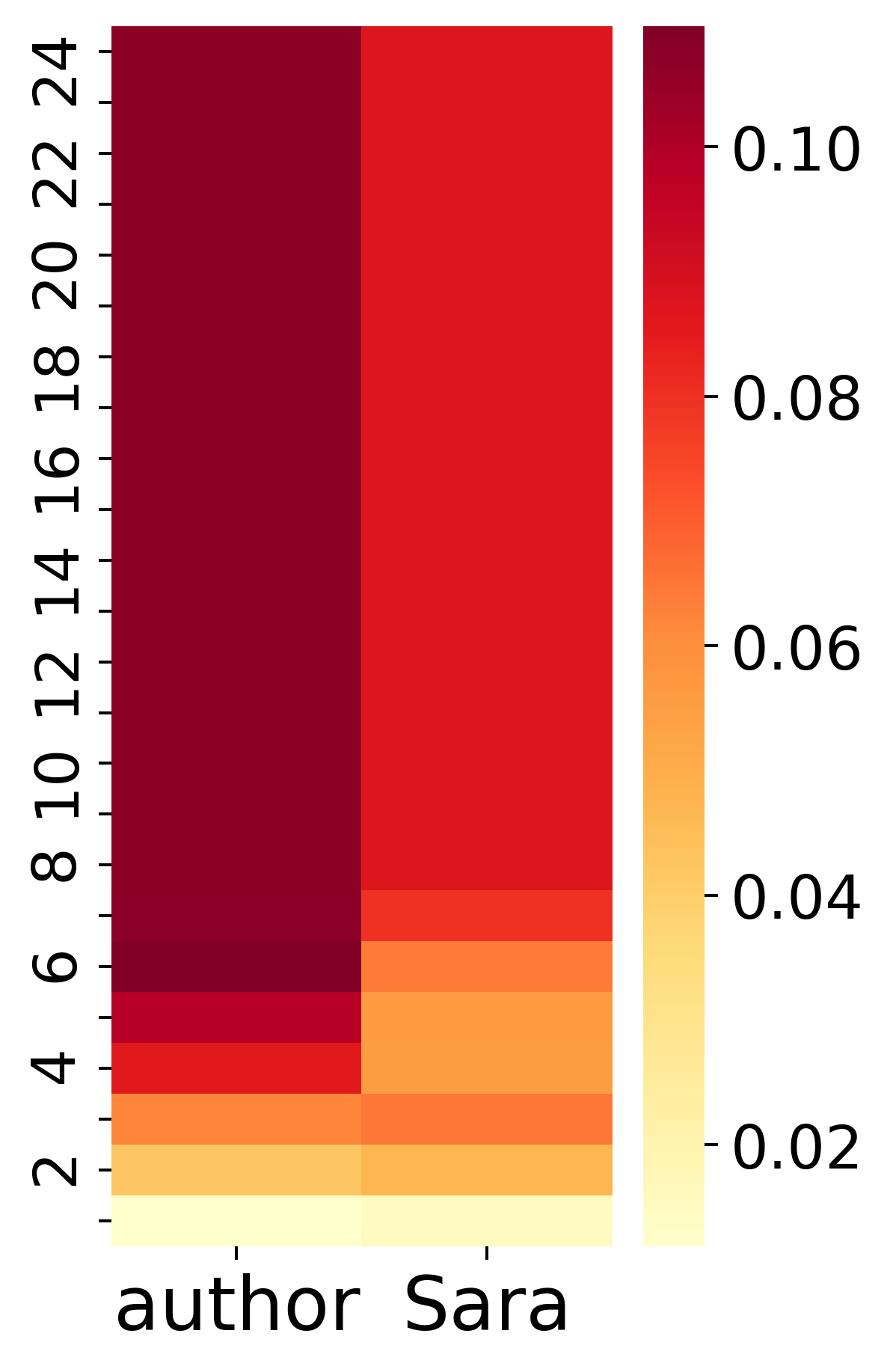}};
    \node[rotate=90] at (-0.1,1.15) {\tiny{\fat}};
     \end{tikzpicture}
     \caption{``\emph{The author talked to Sara about} \texttt{mask} \emph{book}.''
     \label{fig:bert_pronoun_a}}
     \end{subfigure}
     
     \begin{subfigure}[b]{\columnwidth}
     \begin{tikzpicture}
    \node[anchor=south west,inner sep=0] at (0,0)
    {\includegraphics[width=0.118\columnwidth, trim={0.0cm 0.0cm 0.0cm 0.0cm},clip]{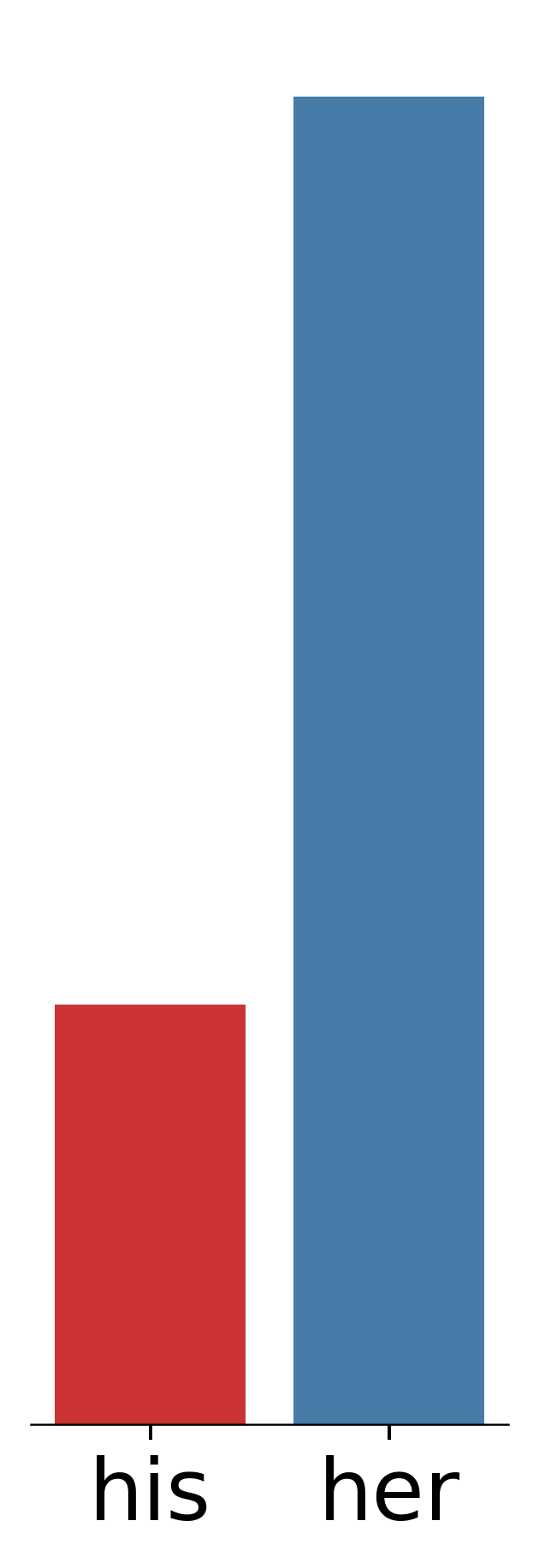}};
    \end{tikzpicture}
         \hfill
    \begin{tikzpicture}
    \node[anchor=south west,inner sep=0] at (0,0)
    {\includegraphics[width=0.22\columnwidth, trim={0.0cm 0.0cm 0.0cm 0.0cm},clip]{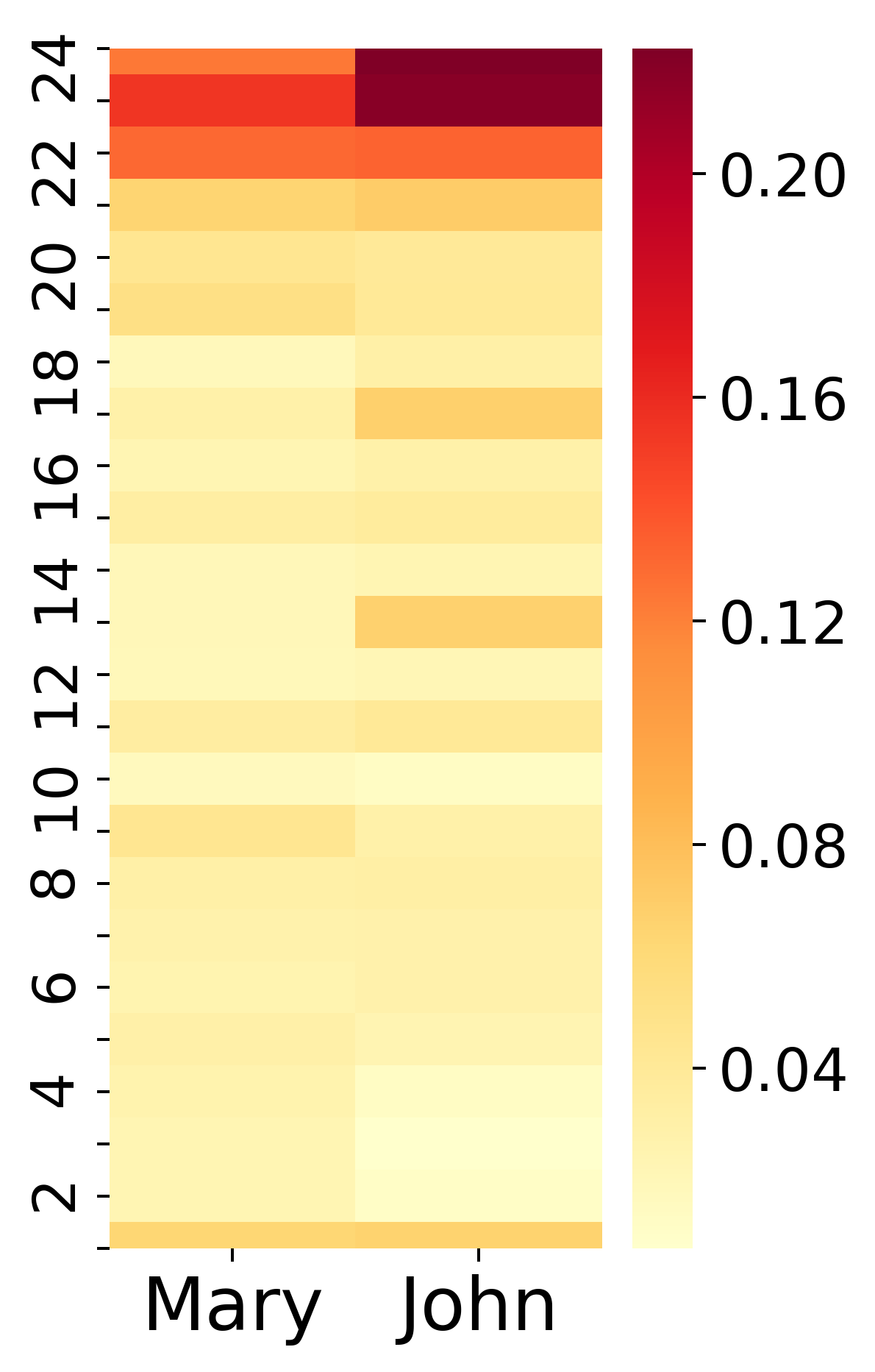}};
    \node[rotate=90] at (-0.1,1.15) {\tiny{\rat}};
     \end{tikzpicture}
          \hfill
    \begin{tikzpicture}
    \node[anchor=south west,inner sep=0] at (0,0)
    {\includegraphics[width=0.22\columnwidth, trim={0.0cm 0.0cm 0.0cm 0.0cm},clip]{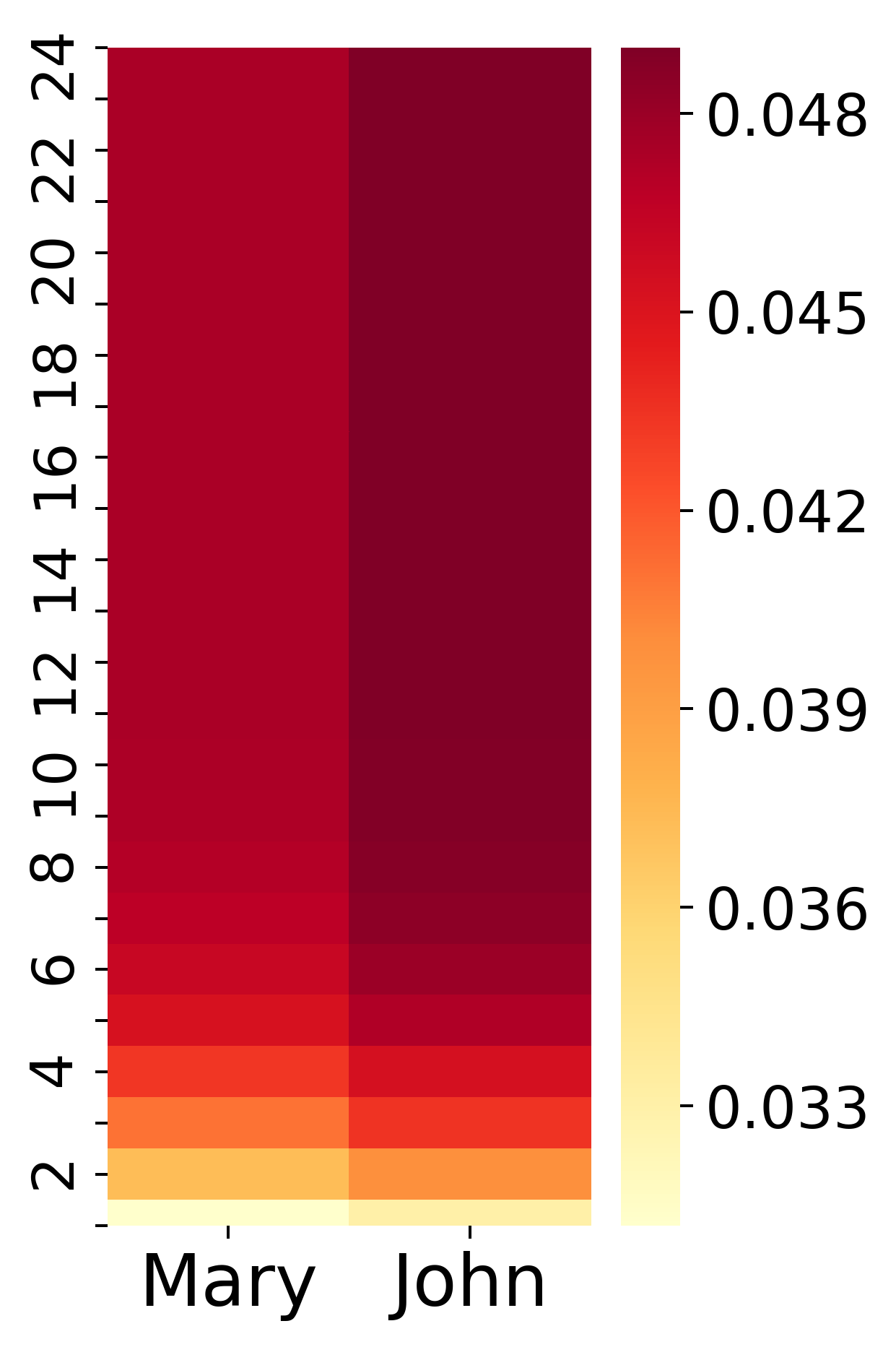}};
    \node[rotate=90] at (-0.1,1.15) {\tiny{\jat}};
     \end{tikzpicture}
          \hfill
    \begin{tikzpicture}
    \node[anchor=south west,inner sep=0] at (0,0)
    {\includegraphics[width=0.22\columnwidth,trim={0.0cm 0.0cm 0.0cm 0.0cm},clip]{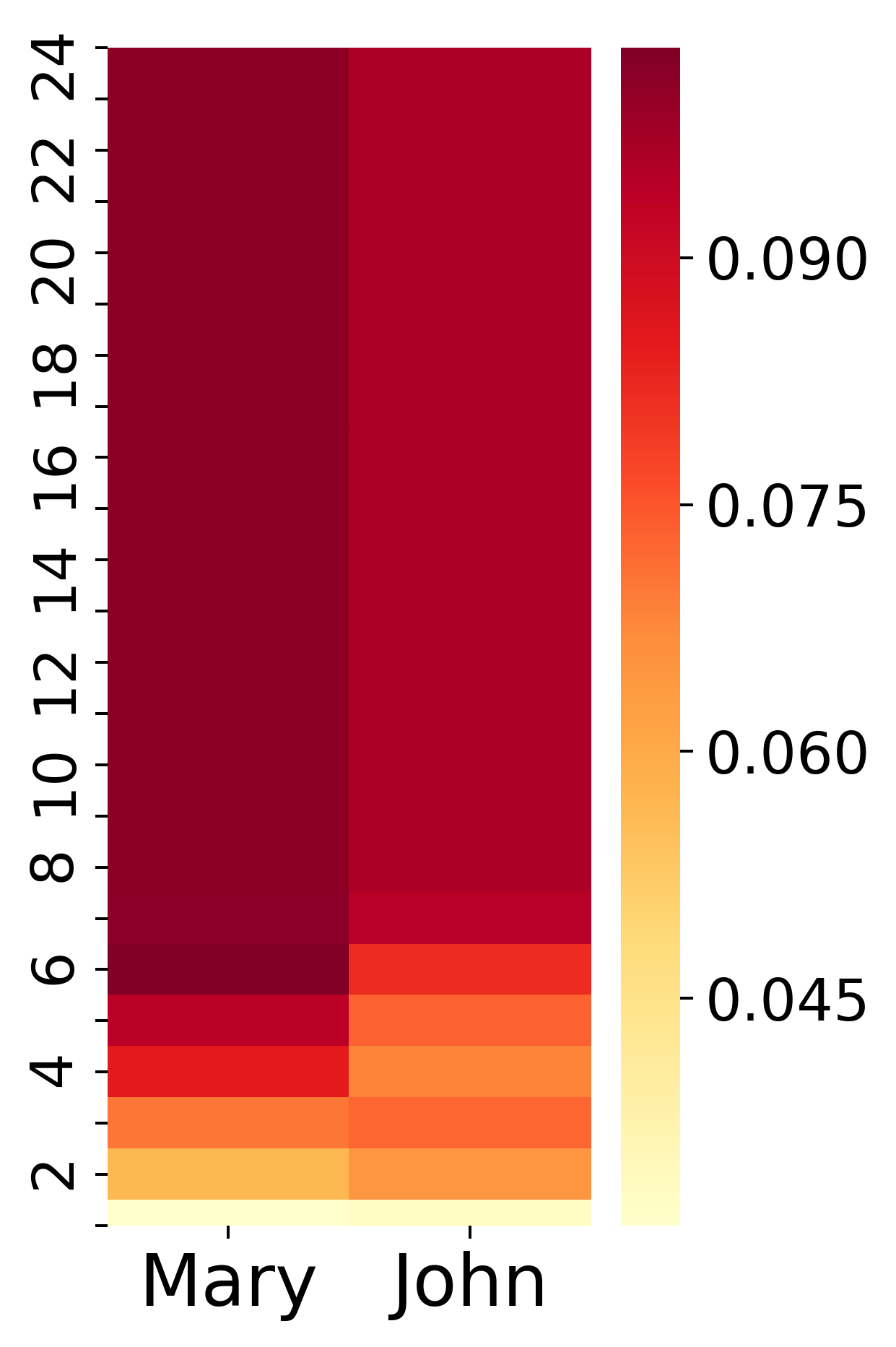}};
    \node[rotate=90] at (-0.1,1.15) {\tiny{\fat}};
     \end{tikzpicture}
     \caption{\emph{``Mary convinced John of} \texttt{mask} \emph{love}.'' \label{fig:bert_pronoun_b}}
     \end{subfigure}
    \caption{Bert attention maps. We look at the attention weights from the \texttt{mask} embedding to the two potential references for it, e.g. ``author'' and ``Sara'' in (a) and ``Mary'' and ``John'' in (b). The bars, at the left, show the relative predicted probability for the two possible pronouns, ``his'' and ``her''.
    \label{fig:bert_pronoun}}
\vspace{-12pt}
\end{figure}

Furthermore, as shown in Table~\ref{tab:quant_comp_blank} and \ref{tab:quant_comp_grad} both \jat and \fat, are better correlated with \blankout scores and input gradients compared to \rat, but \fat weights are more reliable than \jat. 
The difference between these two methods is rooted in their different views of attention weights. 
\Fat views them as capacities, and at every step of the algorithm, it uses as much of the capacity as possible. Hence, \fat computes the maximum possibility of token identities to propagate to the higher layers. Whereas \jat views them as proportion factors and at every step, it allows token identities to be propagated to higher layers exactly based on this proportion factors. 
This makes \jat stricter than \fat, and so we see that \jat provides us with more focused attention patterns. However, since we are making many simplifying assumptions, the strictness of \jat does not lead to more accurate results, and the relaxation of \fat seems to be a useful property.

At last, to illustrate the application of \fat and \jat on different tasks and different models, we examine them on two pretrained BERT models. We use the models available at \url{https://github.com/huggingface/transformers}.

Table~\ref{tab:bertsst_quant} shows the correlation of the importance score obtained from \rat, \jat and \fat from a DistillBERT~\citep{sanh2019distilbert} model fine-tuned to solve ``SST-2''~\citep{socher-etal-2013-recursive}, the sentiment analysis task from the glue benchmark~\citep{wang-etal-2018-glue}. Even though for this model, all three methods have very low correlation with the input gradients, we can still see that \jat and \fat are slightly better than \rat.

Furthermore, in Figure~\ref{fig:bert_pronoun}, we show an example of applying these methods to a pre-trained Bert to see how it resolves the pronouns in a sentence. What we do here is to feed the model with a sentence, masking a pronoun. Next, we look at the prediction of the model for the masked word and compare the probabilities assigned to ``her'' and ``his''. Then we look at \rat, \jat and \fat weights of the embeddings for the masked pronoun at all the layers. 
In the first example, in Figure~\ref{fig:bert_pronoun_a}, \jat and \fat are consistent with each other and the prediction of the model. Whereas, the final layer of \rat does not seem to be consistent with the prediction of the models, and it varies a lot across different layers. 
In the second example, in Figure~\ref{fig:bert_pronoun_b}, only \fat weights are consistent with the prediction of the model.

\section{Conclusion}
Translating embedding attentions to token attentions can provide us with better explanations about models' internals. Yet, we should be cautious about our interpretation of these weights, because, we are making many simplifying assumptions when we approximate information flow in a model with the attention weights.
Our ideas are simple and task/architecture agnostic.
In this paper, we insisted on sticking with simple ideas that only require attention weights and can be easily employed in any task or architecture that uses self-attention.
We should note that all our analysis in this paper is for a Transformer encoder, with no casual masking. Since in Transformer decoder, future tokens are masked, naturally there is more attention toward initial tokens in the input sequence, and both \jat and \fat will be biased toward these tokens. Hence, to apply these methods on a Transformer decoder, we should first normalize based on the receptive field of attention.

Following this work, we can build the attention graph with effective attention weights~\citep{brunner2019validity} instead of raw attentions. Furthermore, we can come up with a new method that adjusts the attention weights using gradient-based attribution methods~\citep{ancona2019}. 

\section*{Acknowledgements}
We thank Mostafa Dehghani, Wilker Aziz, and the anonymous reviewers for their valuable feedback and comments on this work.
The work presented here was funded by the Netherlands Organization for Scientific Research (NWO), through a Gravitation Grant
024.001.006 to the Language in Interaction Consortium.

\bibliography{acl2020}
\bibliographystyle{acl_natbib}

\clearpage

\appendix

\section{Appendices}
\label{sec:appendix}

\subsection{Single Head Analysis}
\label{app:singlehead}
For analysing the attention weights, with multi-head setup, we could either analyze attention heads separately, or we could average all heads and have a single attention graph. However, we should be careful that treating attention heads separately could potentially mean that we are assuming there is no mixing of information between heads, which is not true as we combine information of heads in the position-wise feed-forward network on top of self-attention in a transformer block. It is possible to analyse the role of each head in isolation of all other heads using \jat and \fat. To not make the assumption that there is no mixing of information between heads, for computing the ``input attention'', we will treat all the layers below the layer of interest as single head layers, i.e., we sum the attentions of all heads in the layers below. For example, we can compute \jat for head $k$ at layer $i$ as $\tilde{A}(i,k) = A(i,k) \bar{A}_(i)$, where, $\bar{A}_(i)$ is \jat computed for layer $i$ with the single head assumption.

\end{document}